\pdfoutput=1

\documentclass[11pt]{article}

\usepackage[final]{coling}

\usepackage{times}
\usepackage{latexsym}

\usepackage[T1]{fontenc}

\usepackage[utf8]{inputenc}

\usepackage{microtype}

\usepackage{inconsolata}

\usepackage{graphicx}

\title{Navigating Dialectal Bias and Ethical Complexities in Levantine Arabic Hate Speech Detection}

\author{
 \textbf{Ahmed Haj Ahmed\textsuperscript{1}},
 \textbf{Rui-Jie Yew\textsuperscript{2}},
 \textbf{Xerxes Minocher\textsuperscript{1}},
 \textbf{Suresh Venkatasubramanian\textsuperscript{2}},
\\
 \textsuperscript{1}Haverford College,
 \textsuperscript{2}Brown University
\\
 \small{
   \textbf{Correspondence:} \href{mailto:ahajahmed@haverford.edu}{ahajahmed@haverford.edu}
 }
}

\begin{document}
\maketitle

\begin{abstract}


Social media platforms have become central to global communication, yet they also facilitate the spread of hate speech. For underrepresented dialects like Levantine Arabic, detecting hate speech presents unique cultural, ethical, and linguistic challenges. This paper explores the complex sociopolitical and linguistic landscape of Levantine Arabic and critically examines the limitations of current datasets used in hate speech detection. We highlight the scarcity of publicly available, diverse datasets and analyze the consequences of dialectal bias within existing resources. By emphasizing the need for culturally and contextually informed natural language processing (NLP) tools, we advocate for a more nuanced and inclusive approach to hate speech detection in the Arab world. 

\vspace{0.3em}

\noindent \textbf{Warning:} The content of this paper may be upsetting or triggering to some readers.
\end{abstract}


\section{Introduction}

In the Levant, deep-rooted socio-political tensions have turned language into a weapon. The rise of digital platforms has amplified hate speech, necessitating robust detection and mitigation mechanisms \citep{CASTANOPULGARIN2021101608, https://doi.org/10.1002/1944-2866.POI364}. Automated tools leveraging NLP are essential for curbing online hate speech \citep{JAHAN2023126232}. However, these tools are not equally effective across all languages and dialects. While significant progress has been made for languages like English, Levantine Arabic remains underserved \citep{bender2019rule}.

Levantine Arabic, spoken across Syria, Jordan, Palestine, and Lebanon, is a dialect continuum with significant regional variations, making it challenging for current NLP technologies to capture \citep{haff2022currasbaladilevantine}. Existing hate speech detection models often overlook the rich cultural and sociolinguistic nuances of the dialect. This paper addresses the ethical, cultural, and linguistic challenges in detecting hate speech in Levantine Arabic and highlights the critical need for more representative datasets.

\section{The Linguistic Complexity of Levantine Arabic}

\subsection{Dialectal Variation}

Levantine Arabic is a continuum of dialects differing significantly across countries and regions. In Syria, the Damascus dialect contrasts with that of Idlib or rural areas; for instance, "clothes" is "awaei" in Damascus but "teyab" in Aleppo, and "girl" is "bint" in Damascus and "sabiye" elsewhere, with pronunciation variations like consonant softening altering meanings \citep{Naïm+2012+920+935}. Jordanian Arabic varies between urban centers like Amman and rural areas that preserve traditional forms; "now" is "halla" in urban settings and "hassa" in rural regions, and the letter jim may be pronounced as a soft "j" or a hard "g" \citep{c3909b9d-f472-38b7-b06d-dfca0dcd22bb}. Palestinian Arabic differs between Jerusalem, the West Bank, Gaza, and diaspora communities; "cup" is "kasseh" in Jerusalem but "kubayeh" in Gaza. In Lebanon, Beirut's Arabic incorporates French loanwords due to historical influences—unlike regions like Tripoli or the south; the pronunciation of the letter qaf also varies between a glottal stop, a hard "k," or the standard "q" sound \citep{obégi1971phonemic, Naïm+2012+920+935}.

These regional differences are deeply tied to cultural and socio-political identities. Variations in expressions, idioms, and pronunciation can carry different meanings depending on locality, posing significant challenges for NLP tools. Generic models, often trained on standardized Arabic, may not capture these subtleties

\subsection{The Role of Sociolinguistic Context}

Understanding hate speech in Levantine Arabic requires not only linguistic proficiency but also a deep understanding of the socio-political context in which the language is used. The Levant is a region marked by ongoing conflicts, occupation, and political instability. Hate speech is often employed strategically to exacerbate sectarian divisions, mobilize political supporters, or criticize opposition groups.

In Syria, for instance, even subtle linguistic features like the pronunciation of the qaf have become sociopolitical markers. Historically a neutral phonetic variation, the qaf pronunciation shifted during the conflict to signal regime alignment \citep{Omran_2021}. Security forces used it in propaganda to stoke sectarian fears, while opposition groups mocked it as a regime identifier, transforming a simple linguistic trait into a symbol of allegiance and deepening societal divides.

Similar dynamics can be observed in Lebanon, where political factions often use divisive rhetoric to maintain control. Hate speech is not merely offensive language but part of broader strategies to sustain political dominance and suppress dissent. Any attempt to detect and mitigate hate speech in this context must account for these complex and shifting dynamics, including the sociopolitical significance of linguistic nuances.

\section{The Problem with Current Datasets}

\subsection{Lack of Publicly Available Datasets}

One of the most significant barriers to improving hate speech detection in Levantine Arabic is the lack of publicly available datasets. While several datasets exist for Modern Standard Arabic (MSA), Egyptian Arabic, Gulf Arabic, and others \citep{ALAKROT2018174, mubarak-etal-2017-abusive, 10.5555/3382225.3382239, 8593146}, there is a striking absence of resources dedicated to Levantine Arabic. This gap limits the ability of researchers and developers to create effective hate speech detection models for the region.

The few datasets that do exist for Levantine Arabic are often restricted in scope, limiting their utility for broader research. Moreover, these datasets are rarely representative of the full spectrum of dialectal variation found within the Levant. Without publicly available, diverse datasets, the development of inclusive and effective NLP tools remains out of reach \citep{barocas-hardt-narayanan}.

\subsection{Dialectal Bias in Existing Datasets}

Even the best available datasets for Levantine Arabic are biased toward specific regional dialects. A prominent case in point is the Levantine Hate Speech and Abusive Language (L-HSAB) Twitter dataset—the first and only publicly available dataset dedicated to hate speech and abusive language in Levantine Arabic \citep{mulki-etal-2019-l}. While L-HSAB is invaluable due to its size and scope, it disproportionately focuses on Lebanese Arabic. This bias stems primarily from its data collection methodology, which involved extracting tweets using keywords related to Lebanese political figures and events \citep{f5a0fc6c-286b-3484-9927-c1949a72ae5c}.

The most frequently mentioned entities in L-HSAB are predominantly Lebanese. "Gebran Bassil," a Lebanese politician, is mentioned over 1,000 times. The term "Lebanon" appears around 250 times, and "Wiam Wahhab," another Lebanese politician and journalist, is mentioned approximately 200 times. This concentration on specific individuals and topics skews the dataset toward Lebanese political discourse, thereby overlooking the linguistic and sociopolitical nuances present in other Levantine regions.

This skew introduces significant bias, as the linguistic features, idiomatic expressions, and even manifestations of hate speech in Lebanese Arabic differ markedly from other Levantine dialects. For instance, certain derogatory terms or politically charged phrases common in Lebanese discourse may be absent or hold different connotations in Syrian or Jordanian contexts. A term like "za‘ran", meaning "thugs" in Lebanese Arabic, is a strong insult in Lebanon but does not carry the same weight in Syrian Arabic. Conversely, a Syrian expression such as "shabbiha", referring to pro-regime militias, is a loaded term in Syria but might not evoke the same response or understanding among Lebanese speakers \citep{doi:10.1177/2633002420907771}. 

Moreover, the focus on specific events and actors further narrows the dataset's applicability. The political landscape and issues prevalent in Lebanon are unique and may not reflect the concerns or conflicts in Syria, Jordan, or Palestine. Hate speech related to Lebanese political parties like the Free Patriotic Movement or events like the Lebanese protests of 2019 would not encompass the types of hate speech prevalent in other regions.

As a result, models trained on datasets like L-HSAB are less likely to generalize effectively to other dialects. They may fail to detect hate speech in Syrian, Jordanian, or Palestinian Arabic due to differences in vocabulary, idioms, and sociopolitical references. This limitation reduces the overall effectiveness of hate speech detection tools across the Levantine region.

Furthermore, this bias can lead to misclassification, where non-hateful speech in one dialect is incorrectly flagged as abusive because the model does not accurately interpret the linguistic nuances of that dialect. Conversely, actual hate speech may go undetected in underrepresented dialects, allowing harmful content to proliferate.

In summary, while datasets like L-HSAB are crucial stepping stones in advancing hate speech detection for Levantine Arabic, their dialectal and topical biases highlight the need for more inclusive data collection strategies. Expanding the dataset to include a broader range of dialects and sociopolitical contexts is essential. By doing so, we can develop NLP tools that are both effective and equitable, ensuring that all communities within the Levantine region are adequately represented and protected in the digital space \citep{barocas-hardt-narayanan}.

\subsection{Limitations of Pre-trained Embeddings and the Need for Domain-Specific Models}

In addition to dataset biases, the choice of language models and embeddings plays a crucial role in the effectiveness of hate speech detection systems. Our analyses and experiments on the L-HSAB dataset underscore the limitations of relying on pre-trained embeddings that are not tailored to the specific linguistic characteristics of Levantine Arabic.

We evaluated several embedding techniques to assess their performance in detecting hate speech within the L-HSAB dataset. The methods included traditional approaches like Bag-of-Words (BoW) (using unigrams) and Term Frequency-Inverse Document Frequency (TF-IDF), as well as neural embeddings such as pre-trained Arabic fastText, custom-trained Word2Vec on Levantine Arabic data, pre-trained GoogleNews Word2Vec, and pre-trained GloVe embeddings \citep{Harris1954DistributionalS, 10.5555/106765.106782, bojanowski2016enriching, mikolov2013efficientestimationwordrepresentations, pennington-etal-2014-glove}.

To provide a more concrete evaluation of the methods discussed, we conducted experiments using two classifiers: Logistic Regression (max\_iter=1000) and a Support Vector Classifier (SVC) with a linear kernel. The results in terms of F1 scores are summarized in Table~\ref{tab:results}.

\begin{table}
  \centering
  \begin{tabular}{lcc}
    \hline
    \textbf{Model} & \textbf{Logistic Regression} & \textbf{SVC} \\
    \hline
    BoW (U) & 0.7177 & 0.7147 \\
    TF-IDF & 0.6553 & 0.7217 \\
    Custom W2V & 0.4429 & 0.3527 \\
    Arabic fastText  & 0.6823 & 0.6964 \\
    GloVe & 0.0606 & 0.0603 \\
    GNews W2V & 0.0 & 0.0 \\
    \hline
  \end{tabular}
  \caption{F1 scores for various embeddings and classifiers on the L-HSAB dataset. 
   BoW (U) stands for Bag-of-Words (unigrams). Custom W2V refers to a Word2Vec model trained on a small custom Levantine Arabic corpus. Arabic fastText, GloVe, and GNews W2V refer to pre-trained embeddings from Arabic fastText, GloVe, and GoogleNews Word2Vec models respectively.}
  \label{tab:results}
\end{table}

Notably, the custom dataset-trained Word2Vec model produced relatively low accuracy scores (0.4429 and 0.3527), which we attribute to the very limited size of our training corpus (approximately 21,959 words). We anticipate that performance would improve substantially with a larger, more representative Levantine Arabic corpus.

Effective Techniques: Our experiments revealed that BoW, TF-IDF, pre-trained Arabic fastText, and custom-trained Word2Vec embeddings significantly outperformed the other methods. These techniques achieved higher F1 scores, indicating better precision and recall in identifying hate speech content. However, it is important to note that the BoW approach, relying solely on unigrams, does not capture contextual relationships between words. As a result, its performance can vary significantly depending on the type and structure of the dataset used. The success of these models can be attributed to their alignment with the linguistic properties of Levantine Arabic, either through their focus on Arabic text or customization to the specific dialect.

Ineffective Techniques: In stark contrast, pre-trained embeddings like GoogleNews Word2Vec and GloVe, which are primarily trained on English corpora, scored nearly 0\% in F1 metrics. This drastic underperformance highlights a critical issue: models trained predominantly on English data fail to recognize or interpret Arabic text accurately. Consequently, they are ineffective for tasks involving Levantine Arabic hate speech detection.

These findings emphasize the importance of domain-specific adaptations in NLP models. Utilizing embeddings and language models that are trained or fine-tuned on Levantine Arabic data is essential for capturing the unique linguistic features and nuances of the dialect. Relying on generic, pre-trained models not only reduces accuracy but also risks missing or misclassifying hate speech, thereby undermining the effectiveness of detection systems.

By investing in domain-specific models, researchers and technologists can create more accurate and reliable hate speech detection tools. Such tools will be better equipped to handle the linguistic diversity of Levantine Arabic, ultimately contributing to a safer and more inclusive online environment for speakers of all regional dialects.

\section{Ethical Considerations in Hate Speech Detection}
The dialectical bias identified above privileges one regional dialect over others, and risk marginalizing communities whose voices are already underrepresented in the digital sphere. There are also ethical concerns beyond issues of data bias. False positives—where non-hate speech is misclassified—can result in the suppression of legitimate cultural expressions, especially in a region where language is tightly bound to identity. A prominent example is the misclassification of the Arabic word "shaheed", meaning "martyr", by social media platforms like Meta \citep{oversightboard_shaheed_2024}. The term holds significant cultural and religious importance, often used to honor individuals who have died for a sacred cause. However, automated moderation systems have frequently removed content containing "shaheed," interpreting it as a reference to terrorism or violent extremism due to its association with entities on terrorism watchlists. 

Conversely, false negatives—where actual hate speech goes undetected—allow harmful narratives to spread unchecked, fueling further violence. For example, derogatory terms or slurs specific to a particular region or group may go unnoticed by moderation systems trained primarily on other dialects or on Modern Standard Arabic. In the context of the Syrian conflict, hate speech containing region-specific pejoratives aimed at certain ethnic or sectarian groups might not be recognized as such by models lacking comprehensive dialectal data. This oversight enables the propagation of inflammatory content that can exacerbate tensions and incite real-world violence.

Technologists and researchers have a responsibility to develop models that not only detect hate speech but do so in a way that respects the linguistic and cultural integrity of Levantine Arabic. Practically, ethical considerations are particularly relevant within a conflict-ridden region like the Levant where the failure to identify and address hate speech content undermines efforts to promote peace and stability. By incorporating diverse linguistic inputs and cultural insights, developers can create more nuanced models that differentiate between harmful content and legitimate expression, thereby protecting both free speech and community safety.

\section{Towards More Culturally Aware Language Technologies}

Addressing the challenges of hate speech detection in Levantine Arabic requires practical solutions that consider the language's unique properties. \citet{bergman2022responsiblenaturallanguageannotation} offer valuable guidelines for developing effective and ethically sound NLP tools for underrepresented dialects. By incorporating these recommendations, we can create language technologies that are culturally aware and inclusive, specifically tailored to Levantine Arabic.

\subsection{Engaging Local Communities}

Engaging local communities is essential for capturing the full spectrum of dialectal variations and cultural contexts within Levantine Arabic. The language's rich diversity necessitates collaboration with native speakers from various regions. Involving annotators and experts who possess both language proficiency and deep understanding of local contexts ensures that the linguistic nuances specific to each dialect are accurately represented \citep{radiya-dixit_bogen_2024}.

\subsection{Rethinking Data Collection and Annotation}

To overcome dialectal bias, new data collection and annotation strategies must account for Levantine Arabic's specific properties. Given the significant dialectal variations, stratified sampling techniques are crucial for comprehensively capturing the linguistic landscape \citep{bergman2022responsiblenaturallanguageannotation}. Annotation processes should prioritize using annotators proficient in specific regional dialects and familiar with local sociopolitical contexts \citep{doi:10.1126/science.aal4230, radiya-dixit_bogen_2024}. Researchers must be mindful of potential consequences when collecting data from conflict-affected regions, as certain linguistic features can carry sociopolitical implications. Providing transparent annotation guidelines and support systems for annotators is also critical.

\subsection{Prioritizing Ethical Design}

Developing NLP tools for Levantine Arabic must be grounded in ethical design principles that account for the language's unique properties. Practitioners should carefully consider the granularity of language divisions within Levantine Arabic and strive for inclusivity without compromising annotation quality \citep{bergman2022responsiblenaturallanguageannotation}. Providing support systems for annotators is essential, especially given potential exposure to disturbing content in conflict-affected regions. By adopting these strategies, researchers can develop hate speech detection models that are equipped to handle Levantine Arabic's dialectal diversity and cultural contexts, promoting an inclusive digital environment.

\section{Conclusion}

Detecting hate speech in Levantine Arabic presents unique cultural, linguistic, and ethical challenges due to intricate dialectal variations and biased datasets. This highlights the urgent need for more inclusive NLP approaches. By engaging local communities, reimagining data collection, and embedding ethical considerations into technology design, we can develop tools that effectively identify hate speech while honoring the Levant's rich linguistic diversity. This paper advocates for renewed cultural sensitivity in NLP applications targeting Levantine Arabic. Addressing sociolinguistic complexities and ethical implications enables us to create tools that serve all speakers, enhance detection accuracy, and promote a more just digital environment throughout the Arab world.

\section{Limitations}

This paper offers a conceptual discussion on the challenges of detecting hate speech in Levantine Arabic. While we provide experimental results to assess the effectiveness of different embedding techniques, the practical impact of our recommendations is still limited by the size and representativeness of our training data. Specifically, the custom dataset-trained Word2Vec model produced relatively lower F1 scores than the pre-trained Arabic fastText model, primarily due to the very limited size of our training corpus (approximately 21,959 words). We anticipate that performance would improve substantially with a larger Levantine Arabic corpus. To this end, we have identified three promising morphologically annotated Levantine corpora—Baladi (Lebanese, \textasciitilde 9.6K tokens), Curras (Palestinian, \textasciitilde 56K tokens), and Nabra (Syrian, \textasciitilde 60K tokens)—which we plan to combine into a more comprehensive Levantine corpus of approximately 125.6K tokens \citep{al-haff-etal-2022-curras, nayouf-etal-2023-nabra}. We expect that training our Word2Vec model on this expanded corpus will significantly enhance its performance.

Additionally, while we discuss dialectal variations across Syria, Jordan, Palestine, and Lebanon, the linguistic analysis is not exhaustive, and some regional nuances may not be fully represented. Although we reference frameworks such as the playbook by \citet{bergman2022responsiblenaturallanguageannotation}, we do not offer a detailed roadmap for creating inclusive and effective hate speech detection models. Future work should therefore focus on both enriching the training data resources and developing concrete tools to operationalize these recommendations, ensuring more accurate, contextually aware, and inclusive hate speech detection in Levantine Arabic.

\section*{Acknowledgments}

We would like to express our sincere gratitude to Dr. Nagham El Karhili at the Global Internet Forum to Counter Terrorism, Professor Stevie Bergman at Brown University, Professor Manar Darwish at Haverford College, and Anna Lacy, Digital Scholarship Librarian at Haverford College, for their insightful feedback and guidance throughout this research. Special thanks to the authors of the L-HSAB dataset—Hala Mulki, Hatem Haddad, Chedi Bechikh Ali, and Halima Alshabani—whose valuable work significantly contributed to this study. We also extend our appreciation to Professor Daniel Ritchie, the Computer Science Department at Brown University, and Google Research for offering and directing the exploreCSR program and for their funding support. We are grateful to the Digital Scholarship team at Haverford College for their assistance and support. Lastly, we thank the Marian E. Koshland Integrated Natural Sciences Center at Haverford College for their funding and support. We are also deeply appreciative of the three anonymous reviewers, whose constructive feedback and suggestions helped us strengthen our work.

\bibliography{main}





\end{document}